\documentclass[conference]{IEEEtran}
\IEEEoverridecommandlockouts

\usepackage{cite}

\usepackage{amsmath,amssymb,amsfonts}
\usepackage{algorithmic}
\usepackage{algorithm}
\usepackage{graphicx}
\usepackage{textcomp}
\usepackage{xcolor}
\usepackage{microtype}
\usepackage{subfigure}
\usepackage{booktabs}
\usepackage{hyperref}
\usepackage{multirow}

\begin{document}

\title{Energy-efficient Task Adaptation for NLP Edge Inference Leveraging Heterogeneous Memory Architectures}

\author{
\IEEEauthorblockN{Zirui Fu, Aleksandre Avaliani, Marco Donato}
\IEEEauthorblockA{
Tufts University, Medford, MA, USA\\}
}

\maketitle

\begin{abstract}
    Executing machine learning inference tasks on resource-constrained edge devices requires careful hardware-software co-design optimizations. Recent examples have shown how transformer-based deep neural network models such as ALBERT can be used to enable the execution of natural language processing (NLP) inference on mobile systems-on-chip housing custom hardware accelerators. 
    However, while these existing solutions are effective in alleviating the latency, energy, and area costs of running single NLP tasks, achieving multi-task inference requires running computations over multiple variants of the model parameters, which are tailored to each of the targeted tasks. This approach leads to either prohibitive on-chip memory requirements or paying the cost of off-chip memory access. 
    This paper proposes adapter-ALBERT, an efficient model optimization for maximal data reuse across different tasks. 
    The proposed model's performance and robustness to data compression methods are evaluated across several language tasks from the GLUE benchmark. 
    Additionally, we demonstrate the advantage of mapping the model to a heterogeneous on-chip memory architecture by performing simulations on a validated NLP edge accelerator to extrapolate performance, power, and area improvements over the execution of a traditional ALBERT model on the same hardware platform.
\end{abstract}


\section{Introduction}

\begin{figure}[b!]
    \centering
    \includegraphics[width=8.3cm]{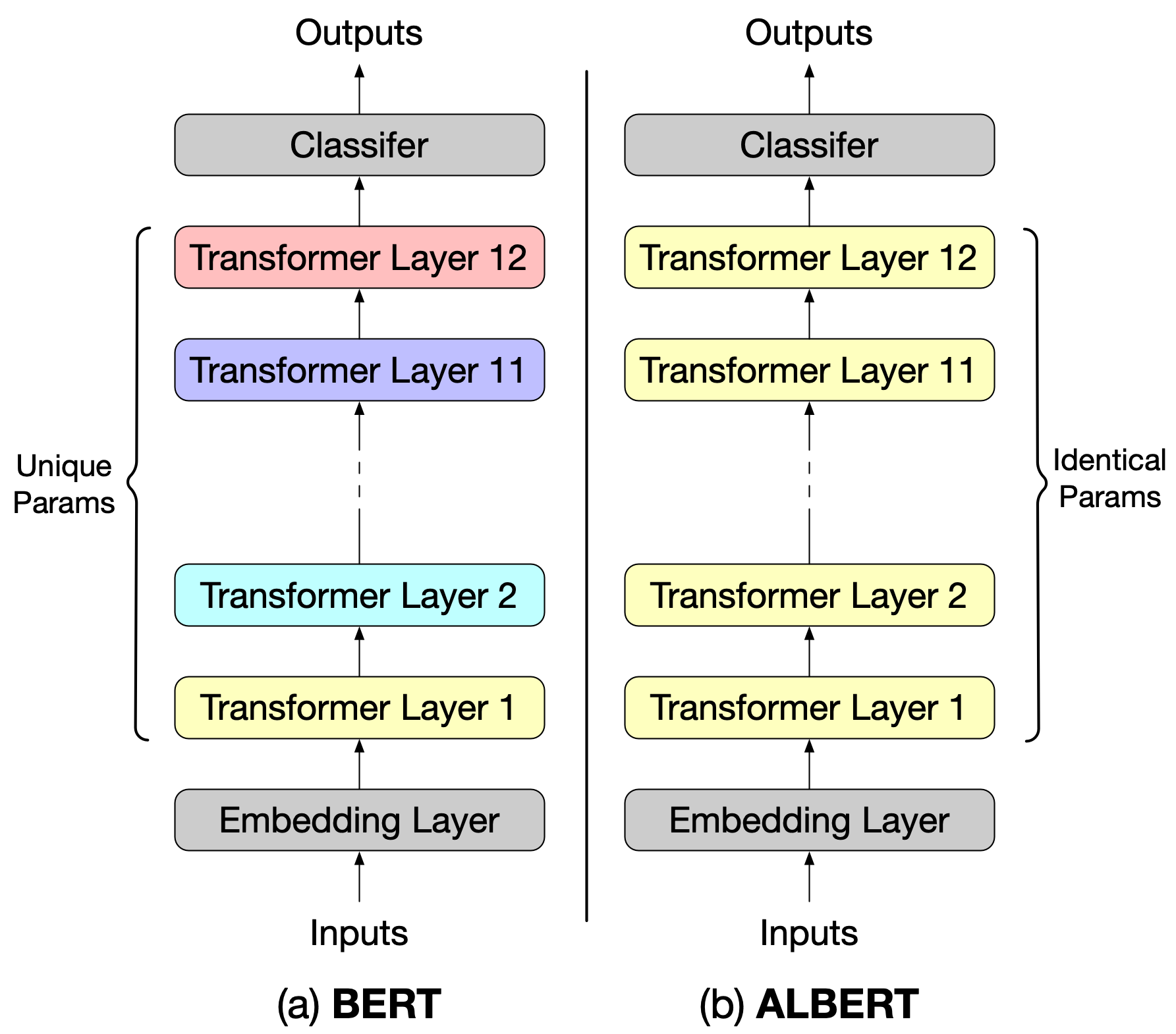}
    \caption{Comparison between BERT (a) and ALBERT (b) models. The BERT model needs to keep twelve unique layers of different parameters. The ALBERT model shares identical parameters across its twelve transformer layers, resulting a significant reduction of numbers of parameters.}
    \label{fig:bert-comp}
\end{figure}

State-of-the-art deep neural network (DNN) large language models (LLMs) are rapidly increasing in size and complexity to embrace emergent abilities and tackle advanced intelligent computations \cite{weiEmergent2022}. This trend poses challenges to power- and resource-limited mobile system-on-chip (SoC) on which the models are stored and executed. In the natural language processing (NLP) domain, for example, after the proposal of the Transformer \cite{vaswaniAttention2017} architecture, the BERT and GPT model families have continuously made breakthroughs in real-world applications. 
The $\mathrm{BERT_{large}}$ model has 340 million parameters \cite{devlinBERT2019} and the GPT-3 model has reached record-breaking 175 billion parameters \cite{brownLanguage2020}, leaving memory footprints of 1.27GB and 652GB using single-precision data representation format correspondingly. Consequently, their enormous memory requirements and data movement costs hinder direct deployment onto mobile SoCs.

Many edge and IoT systems are build in a form factor that blends in with the surrounding environment. For such systems, traditional input modalities are not a viable option. On the other hand, DNN models introducing conversational AI capabilities have the potential to introduce more natural and proactive interactions, especially when running multi-task inference (MTI) operations aimed at combining tasks such as paraphrasing, sentimental analysis, and question-answering. Although existing solutions are effective in alleviating the bottlenecks of running a single NLP task, achieving low-latency MTI requires running computations over multiple variants of the model parameters, which are tailored to each of the targeted tasks. This approach leads to either prohibitive on-chip memory requirements or expensive off-chip memory access. Additionally, the cost of deploying multiple copies of the same model increases with additional tasks.

For example, the ALBERT architecture~\cite{lanALBERT2020} reduces the number of model parameters by 10$\times$ compared to BERT by replacing the twelve distinct Transformer encoder layers with twelve identical layers, achieving a higher-level parameter reuse as shown in the structural comparison in Figure \ref{fig:bert-comp}. Much like its beefier relatives, ALBERT implements a language model that is usually pre-trained on large text corpora to learn both low- and high-level language features and can be used as foundation for transfer learning to easily perform task-switching by fine-tuning the transformer layers. The relatively small footprint and the ability to efficiently learn new tasks through fine-tuning make ALBERT a good candidate for edge deployment. However, while task adaption via fine-tuning is an effective solution to optimize training for a specific task, it suffers from catastrophic forgetting, or losing information on previously learned tasks. This limitation makes the fine-tuning approach inadequate to address many real-world DNN application scenarios that require running inference across a variety of tasks. 
Alternative solutions such as TinyTL \cite{caiTinyTL2021}, MCUNet \cite{linMCUNet2020}, and custom models converted using edge-friendly lightweight libraries~\cite{davidTensorFlow2021}, have been proposed specifically to run on embedded devices with only KBs of memory. However, these solutions are not scalable when the target application requires to run either more complex DNN models or multiple inference modalities. The compressed models also have the downsides of being highly task-specialized, losing generalization capabilities, and requiring considerable training efforts for better accuracy when changing tasks. 

In contrast to these solutions, we offer a hardware and software co-design addressing MTI challenges of LLMs on-chip storage and execution. We first optimize the ALBERT model to perform more efficient task adaptation by integrating residual adapters in the network. Residual adapters were originally introduced to enable efficient learning of new tasks with minimal parameter overhead in ResNet for computer vision tasks~\cite{rebuffiLearning2017, rebuffiEfficient2018}, and their application to NLP was first demonstrated in combination with the BERT model~\cite{houlsbyParameterEfficient2019}. In contrast to the adapter-BERT model, for which the primary goal is to enable efficient training for transfer learning \cite{yosinskiHow2014}, we modify ALBERT to incorporate residual adapters as a way to address MTI applications by maximizing parameter reuse while minimizing the memory footprint.
The resulting \textbf{adapter-ALBERT} model inherits the majority of its parameters from a pre-trained vanilla ALBERT model as \textbf{backbone layers}, while its \textbf{task-specific layers} can be trained on new inference scenarios.

Our approach of partitioning the model between task-agnostic parameters in the backbone layers and task-specific parameters in the non-fixed layers, provides a viable pathway to both efficiently store the entire set of parameters for a single task on chip, and reduces the costs of DRAM memory access by requiring to update only a small number of task-specific parameters for executing new tasks.
However, the modifications introduced by the adapter-ALBERT model alone are not sufficient without the appropriate hardware support.
Therefore, we explore the design of an heterogeneous embedded scratchpad memory architecture in which two technologies are combined: CMOS-based SRAM and non-volatile resistive RAM (RRAM). 
RRAM, like other embedded non-volatile memories~\cite{chenReview2016}, provides high memory density, low energy consumption, and comparable read latency when compared to SRAM. However, these emerging devices, while promising, have not yet reached the maturity to fully replace SRAM because of their underwhelming write performance, limited endurance, and lower reliability. These factors make non-volatile memories, including RRAM, a uncompelling for workloads displaying frequent write operations. 
By combining these two technologies, we take advantage of the high storage density and non-volatility in RRAM and mask its non-idealities by using this portion of the heterogeneous memory to store the backbone layers exclusively. On the other hand, non-fixed layers which are more susceptible to updates at runtime are stored in SRAM with minimal added cost.

A similar approach was introduced in MEMTI~\cite{donatoMEMTI2019} to implement a weight scratchpad for the NVIDIA Deep Learning Accelerator (NVDLA)~\cite{zhouResearch2018} to enable the concurrent operation on several computer vision inference tasks. In a similar fashion, our memory architecture is designed as a dedicated scratchpad memory to support the operation of a validated NLP hardware accelerator EdgeBERT~\cite{tambeEdgeBERT2021a}.

Following the modifications we introduce with the adapter-ALBERT model and its mapping to the heterogeneous scratchpad memory architecture, we perform co-design optimizations including pruning, adapter-size accuracy recovery, quantization, and using a mixture of single-level cell (SLC) and 2-bit multi-level cell (MLC) RRAM to efficiently store bit-mask encoded backbone layers parameters. To the best of our knowledge, these techniques allow us to offer the best performance and energy per inference for NLP multi-task inference running on edge devices. In summary, this paper introduces the following contributions:
\begin{itemize}
    \item We propose a MTI-efficient adapter-ALBERT model that enjoys maximum data reuse and small parameter overhead for multiple tasks while maintaining comparable performance than other similar and base models.
    \item We further optimize the model for deployment in resource-constrained edge devices by running a sensitivity study to evaluate the impact of model compression methods to inference accuracy and multi-task adaptability.
    \item We evaluate the performance of the model when running on an edge system by integrating our proposed heterogeneous embedded memory architecture with an NLP accelerator modeled after EdgeBERT, and report energy consumption, latency, and area for the resulting architecture.
\end{itemize}
\section{Model and Memory Architectures}

\begin{figure*}[ht]
    \centering
    \includegraphics[width=\textwidth]{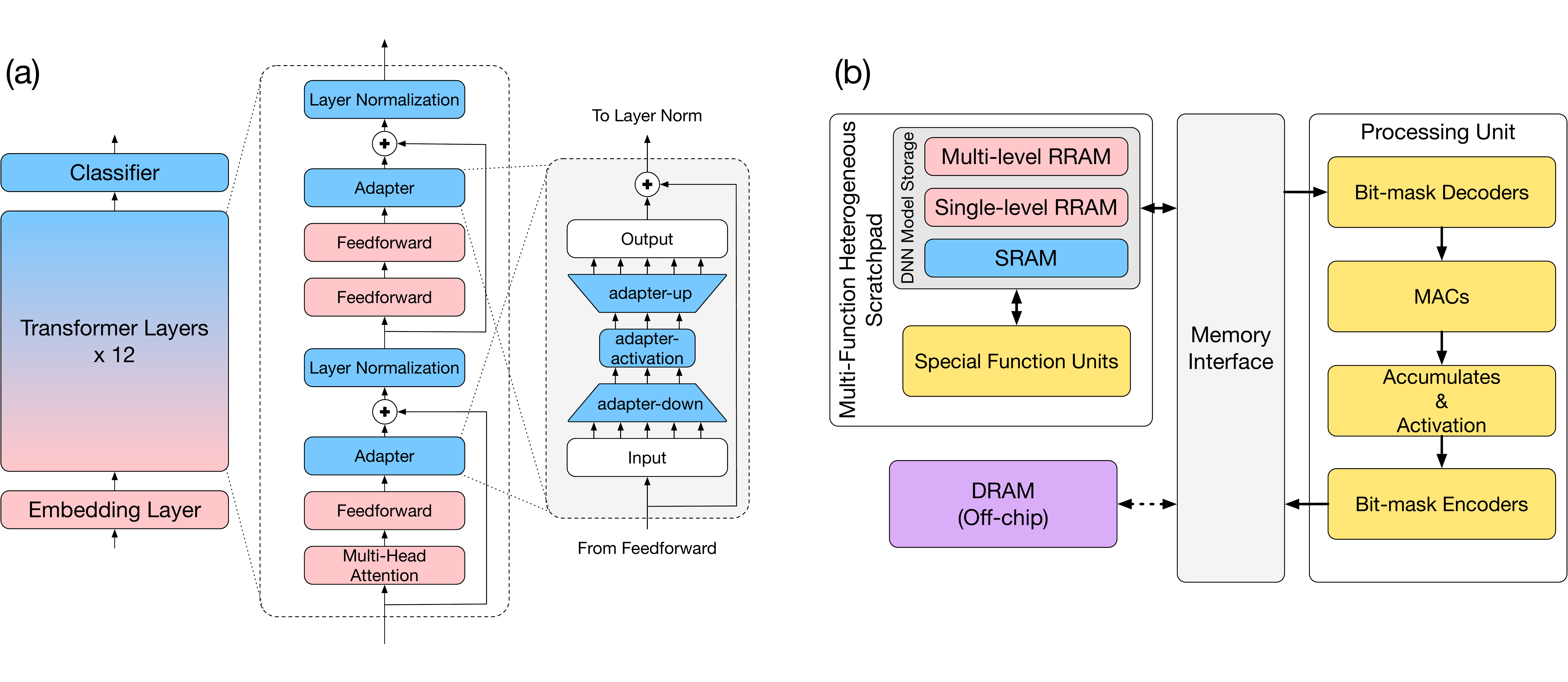}
    \caption{The overview of the adapter-ALBERT model (a) and the HMA (b) architectures. The colors of the adapter-ALBERT model indicate the backbone layers (red) and non-fixed layers (blue). The colors of the HMA architecture indicate different roles of components: red and blue are HMA memory blocks and their colors match the parts of the adapter-ALBERT model; Purple DRAM block is off-chip memory; Yellow blocks belong to the EdgeBERT accelerator.}
    \label{fig:adapter-albert}
\end{figure*}

\subsection{The Adapter-ALBERT Model Architecture}
The block diagram of Figure \ref{fig:adapter-albert}(a) shows the structure of the adapter-ALBERT model. At a high level, the model is based on the same Embedding-Transformer-Classifier skeleton as the vanilla ALBERT model. The blue and red colors are used to differentiate between task-specific and backbone layers, respectively. The embedding layer, highlighted in red, can be fully shared across a variety of NLP tasks, while the classifier, highlighted in blue, is specifically trained on the task at hand by design. For traditional BERT-based language models, the transformer layers would need to be entirely fine-tuned to learn new tasks. However, the transformer layers in adapter-ALBERT are defined by a mixture of trainable and fixed layers. 
The transformer layer partition is shown in greater detail in the inset flow diagram in the middle of Figure \ref{fig:adapter-albert}(a). Going through the diagram from bottom to top, the multi-head attention layer and all feed-forward layers are fixed and use parameters inherited from a pre-trained model, while layer normalization parameters and the newly introduced adapters are considered task-specific layers. In particular, layer normalization parameters are made re-trainable to ensure correct normalization of current data, preventing unmatched data alignments to damage the model performance.
The introduction of adapter blocks, allows the model to learn new features at the cost of small parameter overhead.  
The two residual adapter modules have an identical structure as shown on the right end of the flow diagram.
An adapter module consists of an adapter-down layer that reduces the input feature size down to the adapter bottleneck size, an adapter-activation layer, and an adapter-up layer that increases the feature size back up to the output layer size before being added to the input feature values through a shortcut connection.
The input size of the adapter-down layer and output size of the adapter-up layer match the hidden feature size of 768 in the ALBERT model. The bottleneck size is defined in our model as a hyperparameter, and it plays an important role in determining the model inference accuracy. 
In particular, we will show that distinct adapter blocks can use different bottleneck sizes to prevent accuracy loss.

\subsection{The Heterogeneous Memory Architecture}
The block diagram of Figure \ref{fig:adapter-albert}(b) shows the architecture of the targeted edge system, with emphasis on the heterogeneous scratchpad memory. 
We use the same color coding to describe how task-specific and backbone parameters for the adapter-ALBERT model are stored on chip. 
The SRAM memory is used to store both activations and task-specific parameters. In addition, two separate RRAM memory blocks are used for storing the backbone parameters in sparse format using bitmask encoding. We perform bitmask encoding after pruning both the embedding and backbone transformer layers. The resulting sparse matrices are represented by a vector of non-zero values and a corresponding bitmask array to encode their location. 
Many non-volatile memories, including RRAM, are capable of storing multiple bits in a single memory cell. The multi-level cell (MLC) feature in RRAM is often used to increase storage density. However, MLC RRAM suffers from reduced read margins between adjacent resistance levels and therefore is subject to higher bit error rates. While DNN applications have a demonstrated resilience to fluctuations in the value of the model weights, even single bit-flips in the bitmask could have catastrophic effects on the model accuracy. Error correcting codes (ECCs) are a common approach to make fault-prone memories more robust, but they introduce additional complexity in terms of encoding and decoding data stored in memory, which would quash the benefits of implementing a simple sparse encoding scheme. Therefore, we limit the storage of non-zero values to MLC RRAM, while we adopt a more conservative single-level cell (SLC) RRAM array for storing the bitmask arrays. 

The memory architecture described above is designed to support the operation of a NLP hardware accelerator modeled after EdgeBERT~\cite{tambeEdgeBERT2021a}. The main computational units in the accelerator are depicted in the block diagram with the yellow blocks.
The processing unit contains bitmask encoders and decoders, multiply-and-accumulate (MAC) units, and activation units.
Following the original design for the accelerator, we consider a datapath with 256 MAC units operating at a clock frequency of 1 GHz.
The special function units, integrated within the heterogeneous scratchpad memory, are responsible for near-memory data computation, such as element-wise add and layer normalization. 

\begin{table*}
    \caption{Models Performance Comparison on GLUE Datasets}
    \label{tab:model-acc}
    \vspace{-1em}
    \begin{center}
        \begin{tabular}{cc|cccccccccc}
            \toprule
            Model Name & Adapter Size & COLA & MNLI-MM & MRPC & SST-2 & STS-B & QQP & QNLI & RTE & WNLI \\
            \midrule
            BERT            & N/A   & 52.1   & 84.6  & 88.9   & 93.5   & 85.8   & 71.2   & 90.5   & 66.4   & 56.3   \\
            Adapter-BERT    & 64    & 56.9   & 85.3   & 89.6   & 94.2   & 87.3   & 71.8   & 91.4   & 68.8   & 54.9   \\
            ALBERT          & N/A   & 55.6   & 85.4   & 91.6   & 92.8   & 91.1   & 89.4   & 91.3   & 67.9   & 57.8   \\
            Adapter-ALBERT  & 64    & 50.1   & 84.2   & 91.03   & 90.9   & 87.6   & 85.3   & 91.2   & 69.7   & 50.7   \\
            Adapter-ALBERT  & 32 $\sim$ 128 & 51.7   & 84.7   & 90.8   & 91.39   & 87.5   & 85.9   & 91.5   & 71.1   & 54.9   \\
            \bottomrule
        \end{tabular}
    \end{center}
    \vspace{-1em}
\end{table*}

\section{Model optimizations for efficient adaptation}
\subsection{Model Performance Evaluation}
We evaluate the accuracy performance of our adapter-ALBERT model against three alternatives from the same family, namely BERT \cite{devlinBERT2019}, adapter-BERT \cite{houlsbyParameterEfficient2019}, and ALBERT \cite{lanALBERT2020}, on the GLUE benchmark~\cite{wangGLUE2019}. As a first step, we perform a hyperparameter search across the different GLUE datasets. We begin by sweeping the learning rate selecting between 5E-4 and 5E-5, and consider training epochs from 5 to 10. Training and evaluation batch size are set both to 64. This initial set of experiments is performed in two phases: In the first phase, BERT and ALBERT are fine-tuned for each of the GLUE datasets; 
In the second phase, we transfer the backbone layers from BERT and ALBERT to their corresponding adapter-enhanced model and train only the task-specific layers. As mentioned in the previous section, the adapter modules are designed to have their size independently adjusted as an hyperparameter. We show results for both the case in which the adapter size is fixed to 64 across all tasks, and the case in which the value of the adapter size is selected between 32 and 128. The results from this training experiments are summarized in Table~\ref{tab:model-acc}. 
and show that our adapter-ALBERT model is capable of maintaining competitive results to the other three models, and that varying the adapter size can compensate for the lost accuracy. In particular, the adapter-ALBERT model with an adapter size of 64 matches, or in some cases bests, the other three models on MNLI, MRPC, STS-B, QQP, QNLI, and RTE datasets. 
For the rest three datasets, adjusting the adapter size ranging from 32 to 128 helps boosting the performance close to the other models. 

In addition to evaluating accuracy, we also compare the size of the models in terms of number of parameters as it gives a indication of the relative memory footprint for each of these variants (Table \ref{tab:model-size}). The comparison of trainable parameters reflects the cost of re-training the model for new downstream tasks and the storage and movement costs for task-specific parameters under the MTI scenario. 
Besides showing a small parameter overhead compared to their traditional counterpart, the adapter models also have a much lower number of trainable parameters due to the partition between backbone and task-specific layers This distinction is the key feature that allows the model to perform multi-task inference more efficiently: In the event of the edge system changing the performed inference task, the traditional models would require an update on the entire set of trainable parameters while, for adapters, that number of parameters that need to be refreshed is kept to a few percent of the entire model. Note that, even in the case of adopting the largest adapters size, the parameter overhead would not exceed 4$\%$, or 400K parameters.

\subsection{Model Compression}
After identifying the best set of hyperparameters to train the backbone and adapter layers, we focus on evaluating different compression techniques to make the model more suitable for running on edge devices. We evaluate pruning compression techniques that are ultimately combined to minimize the memory footprint for the adapter-ALBERT model.

\subsubsection{\textbf{Cumulative Sparsity Evaluation}}
Pruning is an effective method for compressing a DNN model and remove redundant parameters to reduce computational complexity and improve generalization capability \cite{liangPruning2021}. However, pruning usually causes accuracy degradation that can be recovered when combined with re-training. Previous work has shown that BERT can endure sparsity between 30\% and 40\% with no detrimental effects on the accuracy of either the pre-trained model or the transferred downstream tasks~\cite{gordonCompressing2020}. However, the new learning approach introduced by the adapter modules requires to re-assess the potential impact of pruning on the model's multi-task learning capabilities. Therefore, we conduct a sensitivity study to ascertain the highest sparsity level that can be achieved by adapter-ALBERT. In particular, we designed a series of experiments to identify the critical sparsity point (CSP) as the limit at which pruning causes the accuracy of the model to drop below the un-pruned baseline accuracy. Note that we apply pruning only to the backbone layers,~\textit{i.e.,} embeddings and fixed transformer layers. The task-specific layers represent only a small portion of the entire model and while pruning them would not affect the model size considerably, 
it would trigger dramatic accuracy degradation. 



We conduct several initial experiments that prune and re-train the embedding and transformer layers down to 90\% sparisty. We have summarized two observations from the model during these experiments as a guideline for cumulative sparsity evaluation.

\textit{Observation 1} - Task bias induced in the pre-trained backbone can significantly impact the overall performance. 
It is important to highlight that in our initial evaluation of the adapter models (Table~\ref{tab:model-acc}) we transferred the backbone parameters form the pre-trained vanilla ALBERT model and re-trained only the non-fixed layers. Pruning the model however, requires a different approach since at each incremental pruning step, the model needs to be retrained to recover from the loss of information. Retraining on BookCorpus would be extremely expensive. Therefore, we tested the effect of re-training the model during pruning using one of largest datasets with abundant information among the GLUE datasets, namely MNLI. While this version of the model could noticeably improve the accuracy performance on the MNLI task after pruning, it would also suffer a noticeable accuracy loss of 3\% on QNLI, which can be interpreted as having biased the backbone parameters towards MNLI. 
To minimize this biasing effect on the backbone parameters, we explored an iterative training approach by further fine-tuning the backbone parameters on QNLI and identifying the next dataset with the highest accuracy drop. Proceeding with this approach quickly leads to a loss of generalization capabilities in the backbone layers, especially when required to re-train on the smaller datasets, which exposes the model to overfitting. Therefore, we decided to explore an alternative route to recover the accuracy loss resulting from pruning.

\textit{Observation 2} - Embedding layer and transformer layers show different pruning sensitivities.
In \textit{observation 1}, we analyzed the adapter-ALBERT model's response to a common pruning threshold applied to both embedding and transformer layers. However, a more accurate picture of how adapter-ALBERT reacts to pruning can be drawn by considering independent pruning thresholds for embedding and transformers.
The embedding layer accounts for approximately 39.9\% of the total model parameters, while the transformer layers contain about 60\%. We use cumulative sparsity as a way of normalizing the effect of pruning the individual blocks on the overall sparsity of the entire model. 
The cumulative sparsity is calculated as shown in Equation \ref{eq:1}, with $S_{c}$ representing cumulative sparsity, $S_{embd}$ and $S_{tf}$ denoting the sparsity for embedding and transformer layers, respectively, and $P_{embd}$ and $P_{tf}$ indicating the ratio of the embedding layer and transformer layers parameters to the entire model size.

\begin{equation}
    \label{eq:1}
    S_{c} = S_{embd} \times P_{embd} + S_{tf} \times P_{tf}
\end{equation}
\begin{figure}[t]
    \centering
    \includegraphics[width=0.48\textwidth]{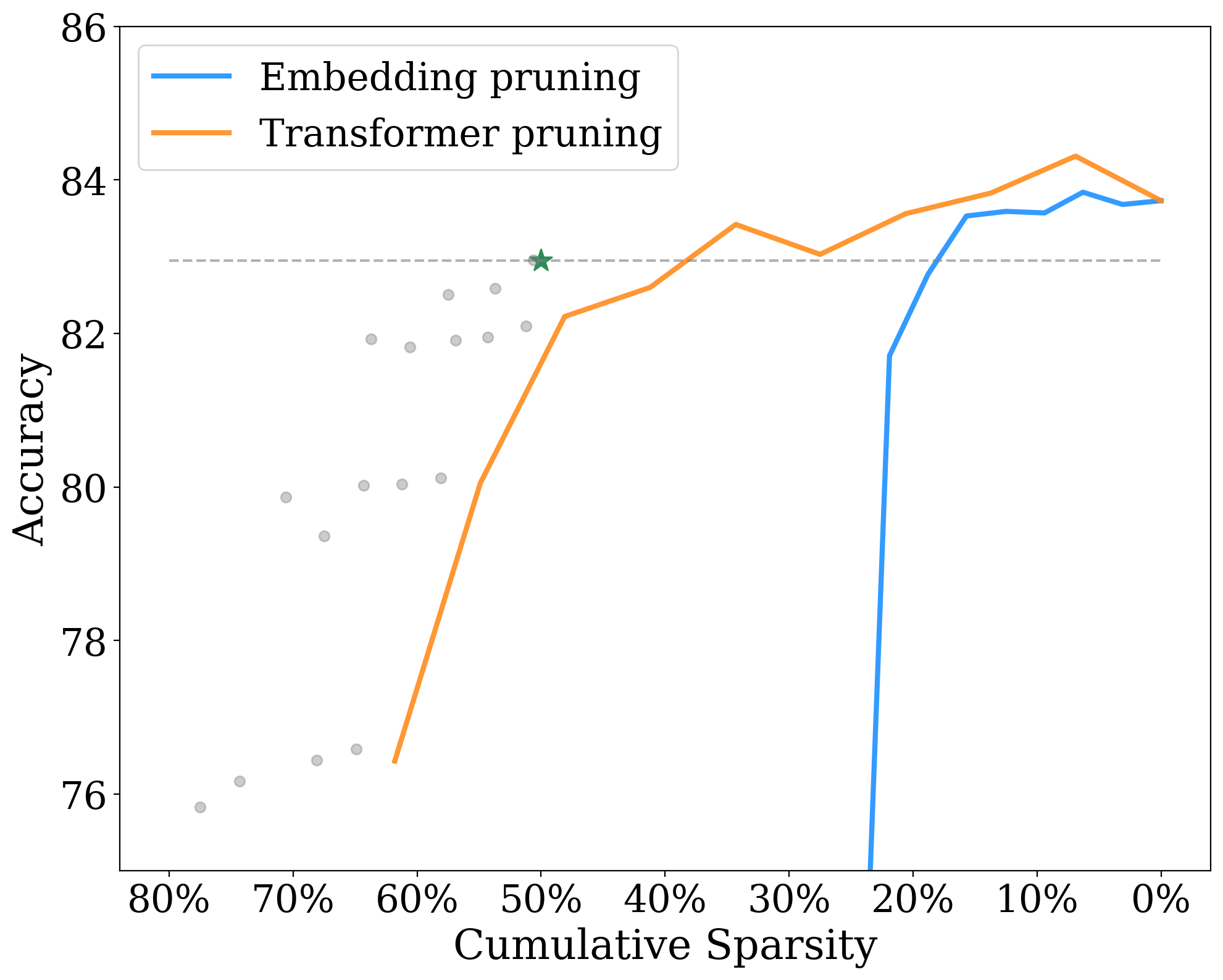}
    \caption{Accuracy comparison for different pruning configurations. The blue and orange lines show the performance trends of the embedded layer and the transformer encoder layers with their isolated pruning cumulative sparsity, respectively. The green star represents the optimal result when combining pruning of both embedding and transformer layers. The gray dots show other pruning combinations with worse accuracy.}
    \label{fig:sparsity}
\end{figure}
\begin{figure*}[ht!]
    \centering
    
\vspace{-1em}
    \includegraphics[width=\textwidth]{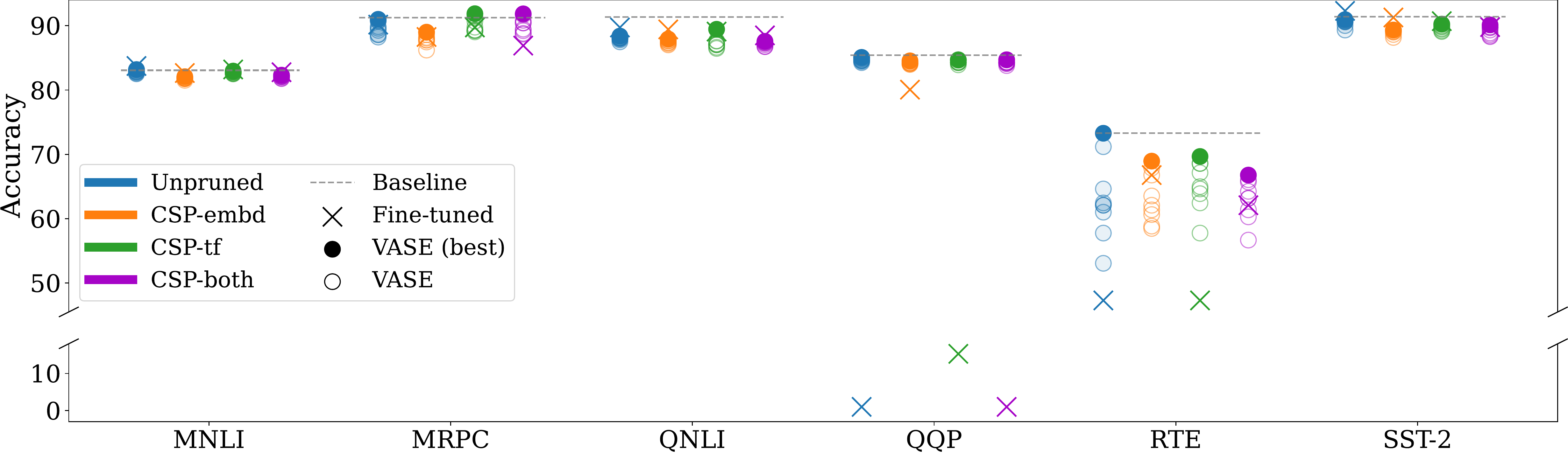}
    \caption{Performance comparison of unpruned model, CSP-embd, CSP-tf, and CSP-all models using fine-tuning and VASE. VASE ($\bigcirc$) outperforms fine-tuning ($\times$) for MRPC, QQP, and RTE, while the two approaches yield similar results for MNLI, QNLI, and SST-2.}
    \label{fig:vase}
\end{figure*}
Based on cumulative sparsity calculated from Eq.~\ref{eq:1}, we show results on accuracy and sparsity level for embedding, transformers, and embedding and transformers together, as summarized in Figure \ref{fig:sparsity}. 
All pruning experiments are executed using the MNLI fine-tuned backbone and adapters.
Pruning only a backbone's embedding layer (blue curve) results in a significant accuracy degradation while providing only moderate sparsity levels around 20\%. These results indicate the high sensitivity of the embedding layer to pruning. 
The transformer layers show a slightly higher resilience to pruning (orange line), with the accuracy dropping below our baseline around 40\% sparsity. This results are in line with findings in literature~\cite{gordonCompressing2020}.
The gray dots show different combinations of embedding-transformer pruning with individual sparsity levels set between 0\% and 90\%. The green star represent the best combination that allows to achieve a cumulative CPS of 50\%, a clear improvement over applying pruning only to embeddings or transformers, which would lead to a CPS of 18.8\% and 41.2\%.
These results indicate that the adapter-ALBERT model is more resilient to embedding layer's sparsity, as the non-fixed layers in the transformer layers are able to learn and compensate these information loss.

\begin{table*}[ht!]
    \caption{Model parameter comparisons in terms of total parameters, trainable parameters, and parameter overhead percentage.}
    \label{tab:model-size}
    \begin{center}
    \begin{small}
    \begin{sc}
        \begin{tabular}{cc|ccc}
            \toprule
            Model Name & Adapter Size & Total Params & Trainable Params & Overhead \\
            \midrule
            BERT            & N/A           & 110M      & 110M      & 0\% \\
            Adapter-BERT    & 64            & 111.2M    & 2.4M      & 1.1\%   \\
            ALBERT          & N/A           & 11.6M     & 11.6M     & 0\% \\
            Adapter-ALBERT  & 64            & 11.88M   & 200.4K     & 2.4\%    \\
            Adapter-ALBERT  & 32 $\sim$ 128 & 11.77M $\sim$ 12.06M   & 100K $\sim$ 400K   & 1.4\%$\sim$3.9\%   \\
            \bottomrule
        \end{tabular}
    \end{sc}
    \end{small}
    \end{center}
\end{table*}

\subsubsection{\textbf{Variable Adapter Size Evaluation}}
While our experiments have shown that the combined pruning of embedding and transformer layers is more conducive to higher CPS values, we still need to verify the multi-task learning capabilities of the pruned backbone. In the next set of experiments, we use four different backbones which we define as follows:
\begin{itemize}
 \item \textbf{Unpruned}: The original un-pruned pre-trained model.
 \item \textbf{CSP-embd}: The model with only embedding layer pruned to the corresponding CSP.
 \item \textbf{CSP-tf}: The model with only transformer layers pruned to the corresponding CSP.
 \item \textbf{CSP-all}:The model with both embedding layer and transformer layers pruned to the corresponding CSPs.
\end{itemize}

We evaluate these backbone variants over six datasets from the GLUE benchmark to highlight three distinct NLP tasks: MRPC and QQP for paraphrasing, SST-2 for sentiment analysis, and MNLI, QNLI, and RTE for natural language inference. To better assist the adapter-ALBERT model in learning this diverse set of tasks we employ a variable adapter size which can be set for each module individually, choosing a value between 32, 64, or 128. This strategy is also compared against the vanilla ALBERT pruned model in which the transformer layers are fine-tuned for each of the sample datasets. The impact of the variable adapter size evaluation (VASE) strategy, can be observed on each model in terms of accuracy performance is presented in Figure \ref{fig:vase}. 

As a general trend, VASE provides significant accuracy improvements on most datasets and the ability to compensate for accuracy loss in pruned models compared to fine-tuning. 


For the unpruned model, almost all of the datasets considered in the evaluation require an adapter size larger than the default setting (64+64), which requires to introduce an additional parameter overhead to preserve inference accuracy.
For the three pruned backbones (CSP-embd, CSP-tf, and CSP-all) the behavior is influenced by the specific task. Nonetheless, we can observe more tasks requiring larger adapter sizes going from CSP-tf to CSP-embd to CSP-all. Although not conclusive, this trend is in line with the growing sparsity level and the need to implement larger adapter sizes to compensate for increasing information loss.



Figure~\ref{fig:vase} shows the comparison of the unpruned and the three pruned backbones trained using VASE against the finetuned version of the model. The baseline thresholds are derived from the results in Table~\ref{tab:model-acc}. The added flexibility in setting different adapter sizes provides a way to combat the backbone bias introduced by fine-tuning the model using the MNLI dataset during the pruning stage. Notably, VASE can recover accuracy loss even in extreme cases such as the finetuned version of QQP, where the accuracy drops below 15\%. Comparing the different backbones in terms of accuracy, CPS-all shows comparable or better results against CPS-embd. CPS-tf has the best accuracy performance across all pruned backbones. As we will show in more detail in the next section, the choice of backbone does not affect the SRAM memory requirements, however selecting CSP-tf over CSP-all would increase the on-chip RRAM capacity by 29.7\%.

\section{Hardware optimizations for edge deployment}



\subsection{Quantization}

While floating point values are still the most common data representation for DNN training, reduced-precision numerical formats are a good compromise for targeting efficient inference~\cite{hubaraBinarized2016,zhuTrained2017,reagenMinerva2016}, since quantized operands can significantly decrease data storage and movement costs, as well as reduce the complexity of the hardware used for implementing MAC processing units~\cite{carmichaelPerformanceEfficiency2019}. 
The vanilla ALBERT model has been demonstrated to benefit from quantization down to 16-bit floating-point without negatively impacting its inference performance~\cite{tambeEdgeBERT2021a}. 
For our adapter-ALBERT model, we hypothesize that, given the adapter module's proven ability to compensate for accuracy loss caused by pruning, it may also exhibit similar trends for quantization.

We have designed experiments with quantization configurations in 16-bit and 8-bit fixed-point representations using the CSP-both backbone. Using the conventional notation, $Q_{i,f}$, to represent the quantization scheme using $i$ bits for integer and sign, and $f$ fractional bits, we focus on $Q_{3, 13}$ and $Q_{3, 5}$.  
To ensure data consistency, the adapter modules are quantized using the same settings as the backbone. The goal for these experiments is to show to what extent the ability of adapter modules in improving or maintaining the inference accuracy is limited by using lower precision quantization. Moreover, we want to verify if adapter modules can learn reduced data information and recover the loss via retraining of the task-specific parameters.

\begin{figure}[t]
    \centering
    \includegraphics[width=8.5cm]{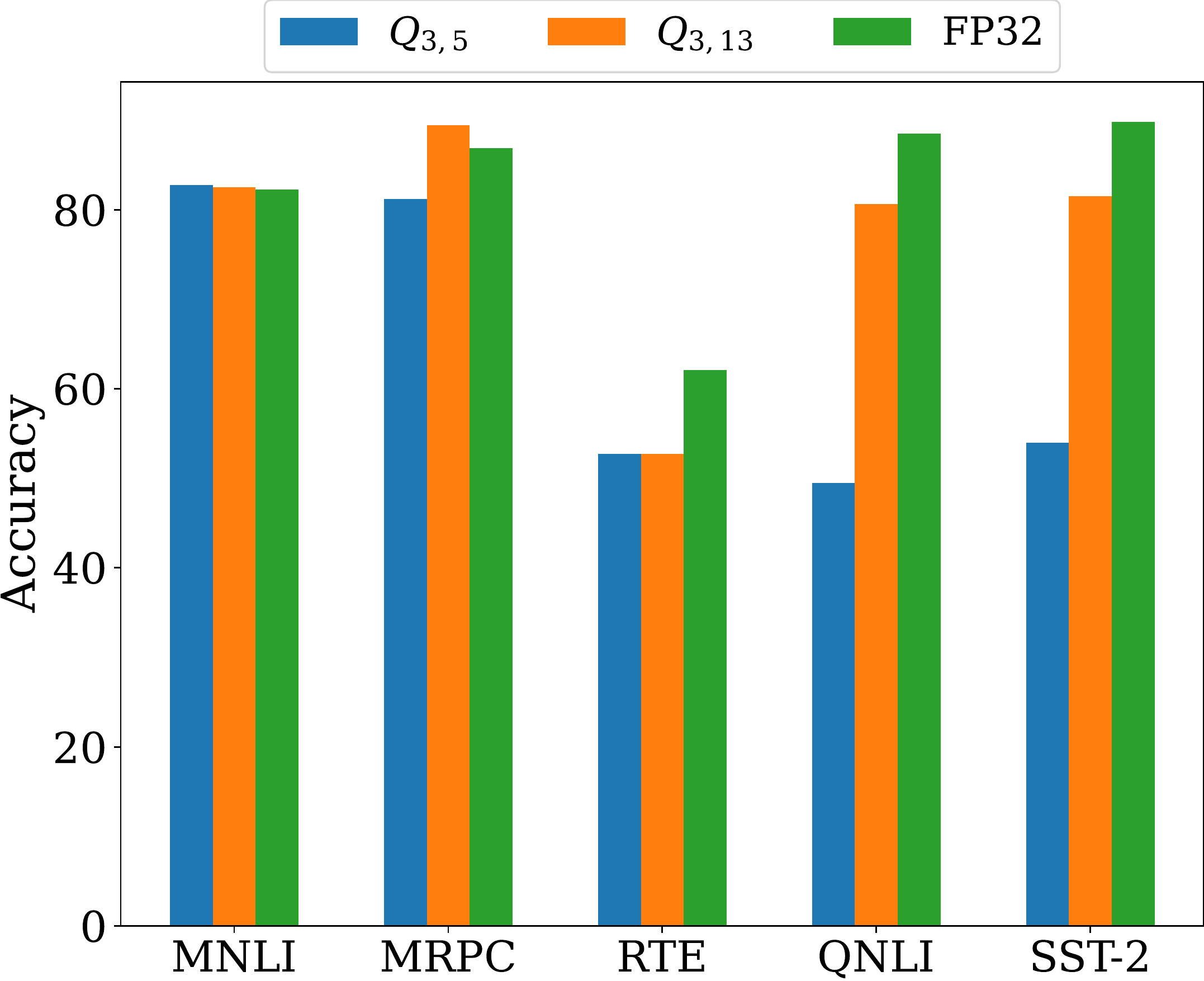}
    \caption{Quantization results on the CSP-all backbone. Although the $Q_{3,13}$ quantization example provides competitive results across all the considered datasets, reducing the number of bits per operand to $Q_{3,5}$ shows a drastic accuracy reduction for some of the QNLI and SST-2 tasks.}
    \label{fig:qt}
\end{figure}

As shown in Figure~\ref{fig:qt}, the $Q_{3,13}$ configuration provides competitive accuracy results when compared with the single-precision floating point (FP32) baseline. 
On the other hand, the results for $Q_{3, 5}$ quantization are much less consistent, suggesting that the optimal quantization scheme will dependent on the subset of tasks used by an application. 

\begin{table*}[ht]
    \caption{Memory Footprint Breakdown for Vanilla ALBERT and Adapter-ALBERT}
    \label{tab:mem-req}
    \begin{center}
    \begin{small}
    \begin{sc}
        \begin{tabular}{c|c|ccc}
            \toprule
            Model & Quant Config & MLC RRAM & SLC RRAM & SRAM  \\
            \midrule
            \multirow{3}{*}{Vanilla ALBERT}  & FP32 & 3.73MB   & 0.47MB & 9.03MB    \\
                                                & $Q_{3, 13}$ & 1.87MB   & 0.47MB & 4.52MB    \\
                                                & $Q_{3, 5}$ & 0.94MB    & 0.47MB & 2.26MB  \\
            \midrule
            \multirow{3}{*}{Adapter-ALBERT}  & FP32 & 11.13MB   & 1.4MB & 0.43MB    \\
                                                & $Q_{3, 13}$ & 5.57MB   & 1.4MB & 0.22MB    \\
                                                & $Q_{3, 5}$ & 2.79MB    & 1.4MB & 0.11MB  \\
            \bottomrule
        \end{tabular}
    \end{sc}
    \end{small}
    \end{center}
\end{table*}

\subsection{Bitmask Encoding}
Leveraging pruning to improve storage density requires an additional step in how the sparse matrices are mapped into the storage system. Several sparse encoding techniques have been proposed in the past and previous work has highlighted how critical it is to guarantee the robustness of these data structures against faults~\cite{pentecostMaxNVM2019}.  
Among the existing techniques, bitmask encoding is a lightweight approach that can be implemented with minimal encoding and decoding hardware overhead. The non-zero values from the sparse matrix are saved in an ordered array and their location is mapped to a binary matrix. At this point that we apply pruning exclusively to the backbone layers. For this reason, it would be advantageous to store the sparse layers in RRAM. While easy to implement, this solution is susceptible to large errors if any of the bits in the bitmask is flipped. To address this issue and preserve the advantage of the density of RRAMs, we split the bitmask and non-zero value data structures to SLC and MLC RRAM arrays respectively. 

\subsection{Accelerator architecture modeling}
To verify the expected improvements introduced by the adapter-ALBERT model optimizations and the associated on-chip memory architecture, we extrapolate the overall system area, energy, and latency by combining results from NVMExplorer~\cite{pentecostNVMExplorer2021} with a performance model tailored around the EdgeBERT accelerator specifications. In order to minimize the on-chip area overhead and be able to deploy our solution onto a mobile SoC, we select a combination of memory macros with different capacities so that the overall memory size is at the closest value exceeding the application requirements. 
When multiple combinations of memory macros are possible, we evaluate each proposal according to their latency and energy consumption. Since macros of different sizes will display different bandwidths, we compute the overall bandwidth as the weighted average of the bandwidth based on the macro's individual capacity. Off-chip memory access to DRAM are modeled using a similar approach to the one introduced in TETRIS~\cite{gaoTETRIS2017}. In our model, we ignore the energy contribution for the computational units, noting that, with the exception of the adapter computations, the two models will perform the same operations.

The baseline for our evaluations is based on the same accelerator and memory architecture, but for the latter, the capacity requirements consider a different model partition in which only the embedding parameters are stored in RRAM, while the rest of the data uses SRAM. As with adapter-ALBERT, the vanilla ALBERT case uses bitmask encoding for the sparse embeddings. The corresponding memory footprint for these two design options under different quantization configurations are shown in Table \ref{tab:mem-req}.
The model parameters partition of adapter-ALBERT introduces higher storage requirements for SLC and MLC RRAM compared to vanilla ALBERT. This is due to the fact that a larger number of parameters are shared across tasks and therefore are kept in the non-volatile portion of the memory. As a consequence, even though we introduce a parameter overhead associated with adapters, the SRAM storage is reduced by a larger factor. The overall effect is that we can take advantage of the denser RRAM storage for a larger portion of the model, leading to a smaller memory footprint.

The workload scenario we consider is that of a MTI application. The on-chip memory stores only the parameters required to process the current task, and the system performs a parameter update from DRAM whenever a new task needs to be executed. Therefore, we provision the on-chip memory footprint based on the worst-case scenario, \textit{i.e.}, the largest set of parameters for any of the given tasks. Note that this corresponds to the dataset with the largest number of classification labels for the vanilla ALBERT case, and in addition,  
must consider the largest adapter size adopted by any of the target tasks when using VASE in the adapter-ALBERT case.

\begin{figure}[ht]
    \centering
    \includegraphics[width=0.48\textwidth]{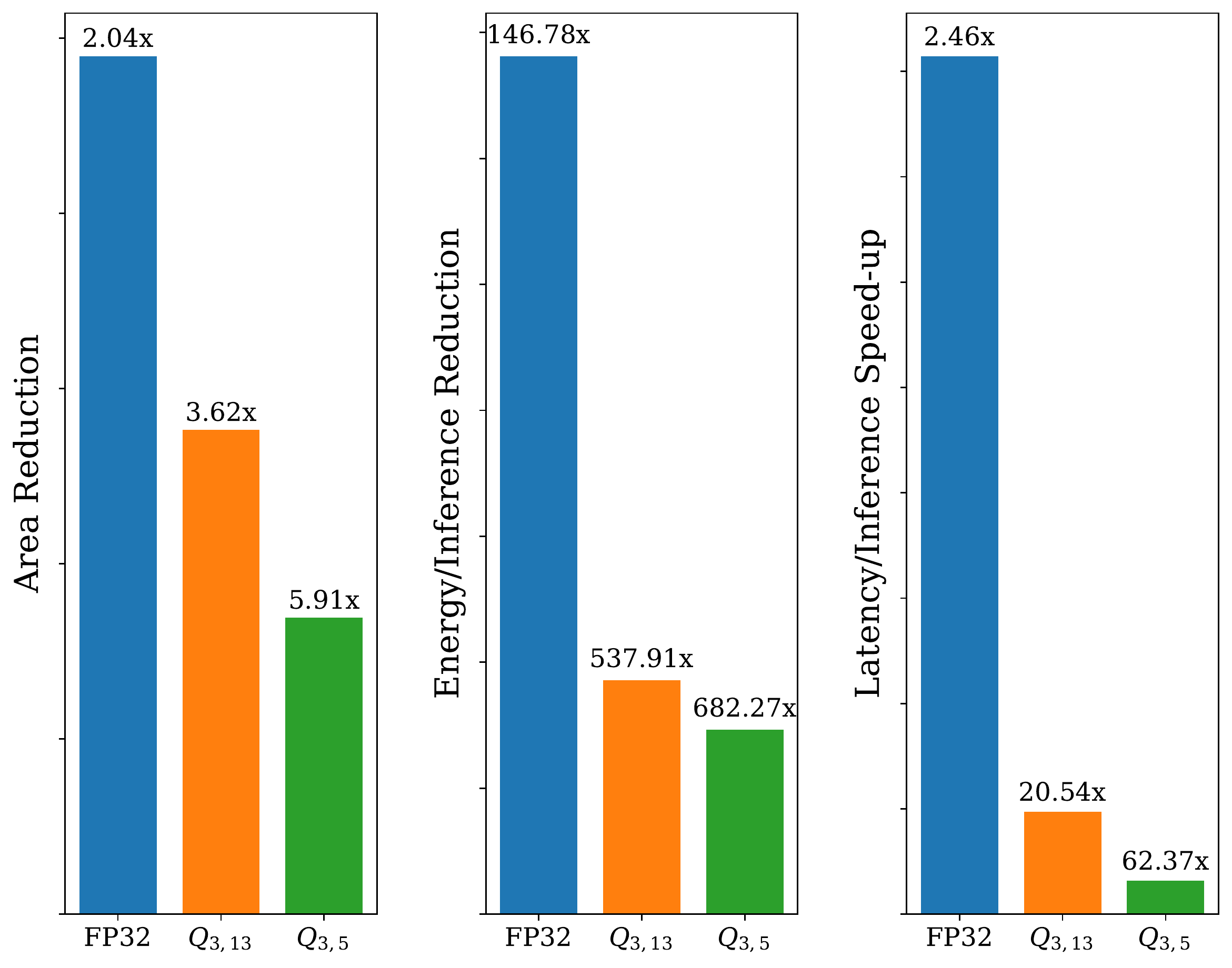}
    \caption{Area, energy/inference, and latency comparison between adapter-ALBERT and vanilla ALBERT using FP32, $Q_{3,13}$, and $Q_{3, 5}$ data types. The results are normalized to the FP32 vanilla ALBERT design.}
    \label{fig:PPA}
\end{figure}

Under this strategy, we compare adapter-ALBERT and vanilla ALBERT's area, energy per inference, and latency per inference under three quantization configurations~\ref{fig:PPA}.
The results are normalized to the FP32 vanilla ALBERT design.
We can observe that for all configurations, the adapter-ALBERT provides all-round advantages against vanilla ALBERT.
Compared to the FP32 implementation under a 3-task MTI scenario, adapter-ALBERT enjoys 2.04$\times$, 146.78$\times$, and 2.46$\times$ reductions in area, energy per inference, and latency per inference. The advantage is even more significant when it comes to 16-bit and 8-bit quantified comparisons. For instance, using the $Q_{3, 5}$ configuration, leads to 5.9$\times$, 682$\times$, and 62$\times$ improvements in area, energy and latency.

\section{Related Work}
\textbf{Multi-Task Learning in Computer Vision -}
Transfer learning DNN models pre-trained on ImageNet~\cite{dengImageNet2009} has been a common approach for efficiently training new tasks in computer vision. However, preserving inference accuracy requires to fine-tune a large fraction of the model. Residual adapter modules~\cite{rebuffiLearning2017, rebuffiEfficient2018} have been introduced as alternative and more efficient way of achieving multi-task learning. The Squeeze-and-Excite blocks proposed by Hu et al.~\cite{huSqueezeandExcitation2019} presents a similar bottle-neck module as residual adapters for channel-wise feature extraction and adaptation. 

Alternative multi-task inference approaches generate multiple copies of the same model which are aggressively compressed to limit the overall model size. Examples of this approach include network slimming~\cite{liuLearning2017} and structured sparsity learning~\cite{wenLearning2016}. The latter approach learns the sparsity from a complex DNN model to accelerate DNN inference. TinyML-inspired projects such as SqueezeNet~\cite{iandolaSqueezeNet2016}, TinyTL~\cite{caiTinyTL2021}, and MCUNet~\cite{linMCUNet2020}, have sought ways to compress the model size to fit in microcontroller-based platforms.

\textbf{Natural Language Processing -}
The field of natural language processing has been rapidly advancing since the proposal of attention mechanism along with the Transformer model~\cite{vaswaniAttention2017}. Edge-cutting attention-based NLP models, including BERT~\cite{devlinBERT2019} and GPT~\cite{radfordImproving, brownLanguage2020}, have brought significant performance improvement over traditional CNN, RNN, and LSTM-based language models and challenges of local deployments due to massive computational and data movement costs from their uncontrollable sizes. Slimmer BERT variants, like ALBERT~\cite{lanALBERT2020}, TinyBERT~\cite{jiaoTinyBERT2020}, and RoBERTa~\cite{liuRoBERTa2019}, have introduced multiple structural and data compression optimizations to reduce their sizes while keeping acceptable performance. Additionally, Gordon et al.~\cite{gordonCompressing2020} have categorized the weight pruning effects on the BERT model, indicating the potential boundary of pruning our adapter-ALBERT model.

Transfer learning is also closely studied in the NLP domain. The adapter-BERT model proposed by Houlsby et al.~\cite{houlsbyParameterEfficient2019} shows feasibility of transplanting adapter modules from CV domain to NLP domain for efficient training compared to traditional fine-tuning methods. Stickland et al.~\cite{sticklandBERT2019} utilize project attention layers across their BERT-PALs model as an alternative task-adaption approach for a lower parameter overhead compared to adapter-BERT model (1.13$\times$ vs 1.3$\times$). However, this model, like many of the alternatives discussed in this section, requires to update all parameters which imposes a heavy toll on the memory bus.

\textbf{Embedded Non-Volatile Memories -}
To match the needs of data-intensive applications, new memory technologies have been proposed since the memristors was first hypotesized in 1971~\cite{chuaMemristorThe1971}. After the first demonstrated physical implementation in 2008~\cite{strukovMissing2008}, the resistive random access memory (RRAM) gradually became one of the most promising eNVM technologies due to its scalability~\cite{dengDesign2013}, storage density~\cite{dengRRAM2013}, competitive read performance~\cite{shyh-shyuansheuFastWrite2011}, and CMOS compatibility~\cite{tanachutiwatFPGA2011}. Collective studies and reviews of RRAM and other eNVM technologies are performed by Panda et al.~\cite{pandaCollective2018} and Chen et al.~\cite{chenReview2016}.

DNN storage architecture realizations utilizing RRAM have been closely studied. Donato et al. have presented an on-chip memory optimization utilizing SRAM and RRAM for CV multi-task inference \cite{donatoMEMTI2019}, which covers the shortcoming of RRAM's write costs by assigning frequently-updated parameters of an adapter-equipped ResNet model onto the SRAM portion of the design. 

\section{Conclusions}
As the deployment of large DNN NLP models on edge devices for  multi-task inference (MTI) workloads becomes more prevalent, there is a critical need to balance on-chip memory demands, accuracy, and reduced data movement costs. The adapter-ALBERT model offers a novel co-design methodology to optimize task adaptation on edge devices.

The proposed adapter-ALBERT model significantly reduces trainable parameters at the cost of minimal parameter overhead, maximizing data reuse and resulting in a more efficient storage plan suitable for MTI scenarios and heterogeneous memory architectures. The model's robustness to various data compression methods, including pruning and quantization, is demonstrated.

By employing a pruning strategy that combines sparsity in both embedding and transformer layers, we explore critical pruning thresholds and show that it is possible to maintain competitive accuracy on the GLUE benchmark with over 50\% cumulative sparsity. We analyze the effects of varying adapter sizes on critically pruned models and conclude that the adapter module can recover the accuracy loss caused by pruning.

Our solution is evaluated on a validated hardware accelerator model which integrates an embedded heterogeneous memory architecture. In its most aggressive optimization, our solution offers almost 6$\times$ area reduction, 682$\times$ energy per inference reduction, and 62$\times$ latency per inference speed-up.

In summary, our proposed adapter-ALBERT model offers a powerful approach for efficient NLP DNN with a low memory footprint, low power requirements, and high accuracy. The model's adaptability to pruning and quantization, and its ability to handle multiple tasks make it a suitable solution for on-chip multi-task inference for resource-constrained edge devices.

\bibliography{main}
\bibliographystyle{hieeetr}

\end{document}